\documentclass[10pt]{article} 

\usepackage[accepted]{rlj} 

\usepackage{amssymb}            
\usepackage{mathtools}          
\usepackage{mathrsfs}           
\usepackage{graphicx}           
\usepackage{subcaption}         
\usepackage[space]{grffile}     
\usepackage{url}                
\usepackage{lipsum}             

\usepackage{mathtools}
\usepackage{amsthm}
\theoremstyle{plain}
\newtheorem{theorem}{Theorem}[section]
\theoremstyle{definition}
\newtheorem{definition}[theorem]{Definition}
\theoremstyle{plain}
\newtheorem*{restatement}{Theorem}

\usepackage{algorithm}
\usepackage{algorithmic}

\title{Maximum-Entropy Exploration with Future State-Action Visitation Measures}

\setrunningtitle{Maximum-Entropy Exploration with Future State-Action Visitation Measures}

\author{Adrien Bolland\textsuperscript{1}, Gaspard Lambrechts\textsuperscript{2,3}, Damien Ernst\textsuperscript{1}}

\emails{adrien.bolland@uliege.be, gaspard.lambrechts@mcgill.ca, dernst@uliege.be}

\affiliations{
$^{1}$\textbf{University of Liège, Belgium} \\
$^{2}$\textbf{McGill University, Canada} \\
$^{3}$\textbf{Mila Québec Artificial Intelligence Institute, Canada}
}

\contribution{
    This paper introduces a new MaxEntRL objective in which the intrinsic rewards that regularize the objective are proportional to the entropy of the discounted distribution of state-action features visited during future time steps; it motivates agents to explore these future features.
    }
    {
    Prior MaxEntRL objectives typically regularize policy entropy or the entropy of the state visitation/occupancy instead.
    }

\contribution{
    The paper shows that the expected sum of these intrinsic rewards is a lower bound on the entropy of the discounted distribution of state-action features visited in trajectories starting from the initial states (equivalently, an alternative MaxEntRL objective).
    }
    {
    This result situates the proposed objective as a surrogate for this alternative MaxEntRL objective and highlights a trade-off between the two.
    }
    
\contribution{
    The paper proves that the conditional feature distribution used in the intrinsic reward is the unique fixed point of a contractive operator, which justifies learning an estimate of this distribution via TD-style bootstrapping with $N$-step transitions and leads to an off-policy estimation scheme for the intrinsic rewards.
    }
    {
    The fixed point and its derivation closely resemble those of successor-style representations of future state visitation \citep{janner2020generative}; this structure can be leveraged for estimating MaxEntRL intrinsic rewards off-policy, which is otherwise often difficult.
    }

\contribution{
    The paper provides an off-policy algorithm for maximizing the new MaxEntRL objective; the reported results highlight a trade-off between within-trajectory and marginal feature visitation, depending on the MaxEntRL objective, and emphasize a step toward improving sample efficient intrinsic reward estimation.
    }
    {
    The best exploration objective to pursue is still an open question despite the vast literature, and sample efficient intrinsic reward estimation is mostly ignored in the literature.
    }


\summary{Maximum entropy reinforcement learning motivates agents to explore states and actions to maximize the entropy of some distribution, typically by providing additional intrinsic rewards proportional to that entropy function. In this paper, we study intrinsic rewards proportional to the entropy of the discounted distribution of state-action features visited during future time steps. This approach is motivated by two results. First, we show that the expected sum of these intrinsic rewards is a lower bound on the entropy of the discounted distribution of state-action features visited in trajectories starting from the initial states, which we relate to an alternative maximum entropy objective. Second, we show that the distribution used in the intrinsic reward definition is the fixed point of a contraction operator and can therefore be estimated off-policy. Experiments highlight that the new objective often leads to improved visitation of features within individual trajectories, in exchange for slightly reduced visitation of features in expectation over different trajectories, as suggested by the lower bound. It also leads to improved convergence speed for learning exploration-only agents. Control performance remains similar across most methods on the considered benchmarks.
}

\begin{document}

\makeCover  
\maketitle  

\begin{abstract}
Maximum entropy reinforcement learning motivates agents to explore states and actions to maximize the entropy of some distribution, typically by providing additional intrinsic rewards proportional to that entropy function. In this paper, we study intrinsic rewards proportional to the entropy of the discounted distribution of state-action features visited during future time steps. This approach is motivated by two results. First, we show that the expected sum of these intrinsic rewards is a lower bound on the entropy of the discounted distribution of state-action features visited in trajectories starting from the initial states, which we relate to an alternative maximum entropy objective. Second, we show that the distribution used in the intrinsic reward definition is the fixed point of a contraction operator and can therefore be estimated off-policy. Experiments highlight that the new objective often leads to improved visitation of features within individual trajectories, in exchange for slightly reduced visitation of features in expectation over different trajectories, as suggested by the lower bound. It also leads to improved convergence speed for learning exploration-only agents. Control performance remains similar across most methods on the considered benchmarks.
\end{abstract}

\section{Introduction} \label{sec:introduction}

Many challenging tasks where an agent makes sequential decisions have been solved with reinforcement learning (RL). Examples range from playing games \citep{mnih2015human, silver2017mastering} to managing energy systems and markets \citep{boukas2021deep, aittahar2014optimal}. In practice, many RL algorithms are applied in combination with an exploration strategy to achieve high-performance control. These exploration strategies usually consist of providing intrinsic reward bonuses to the agent for achieving certain behaviors. Typically, the bonus enforces taking actions that reduce the uncertainty about the environment \citep{pathak2017curiosity, burda2018large, zhang2021noveld}, or actions that enhance the variety of states and actions in trajectories \citep{bellemare2016unifying, lee2019efficient, guo2021geometric, williams1991function, haarnoja2019soft}. In many of the latter methods, the intrinsic reward function is the entropy of some distribution over the state-action space. Optimizing jointly the reward function of the MDP and the intrinsic reward function is called Maximum Entropy RL (MaxEntRL) and was shown to be effective in many problems.

The reward of the MDP was already extended with the entropy of the policy in early algorithms \citep{williams1991function} and was only later called MaxEntRL \citep{ziebart2008maximum, toussaint2009robot}. This particular reward regularization provides substantial improvements in the robustness of the resulting policy \citep{ziebart2010modeling, husain2021regularized, brekelmans2022your} and provides a learning objective function with good smoothness and concavity properties \citep{ahmed2019understanding, bolland2023policy}. Well-known algorithms include soft Q-learning \citep{haarnoja2017reinforcement, schulman2017equivalence} and soft actor-critic \citep{haarnoja2018soft, haarnoja2019soft}. This MaxEntRL objective nevertheless only rewards the randomness of actions and neglects the influences of the policy on the visited states.

In order to enhance exploration, \citet{hazan2019provably} were the first to intrinsically motivate agents to visit different states. It led to maximizing the entropy of the discounted state visitation measure and the stationary state visitation measure. For discrete state-action spaces, optimal exploration policies, which maximize the entropy of these visitation measures, can be computed to near optimality with off-policy tabular model-based RL algorithms \citep{hazan2019provably, mutti2020intrinsically, tiapkin2023fast}. For continuous state-action spaces, alternative methods rely on $k$-nearest neighbors to estimate the density of the visitation measure of states (or features built from the states) and compute the intrinsic rewards, which can afterward be optimized with any RL algorithm \citep{liu2021behavior, yarats2021reinforcement, seo2021state, mutti2021task}. These methods require sampling new trajectories at each iteration; they are on-policy, and estimating the intrinsic reward function is computationally expensive. Some other methods rely on parametric density estimators to reduce the computational complexity and share information across learning steps \citep{lee2019efficient, guo2021geometric, islam2019marginalized, zhang2021exploration}. This additional function is typically learned by maximum likelihood estimation using batches of truncated trajectories. Alternative methods have adapted this MaxEntRL objective to maximize the entropy of states visited in single trajectories \citep{mutti2022importance, jain2024maximum}. When large and/or continuous state and action spaces are involved, relying on parametric function approximators is likely the best choice. Nevertheless, existing algorithms are again on-policy. They require sampling new trajectories from the environment at (nearly) every update of the policy, and cannot be applied using a buffer of arbitrary transitions, in batch-mode RL, or in continuing tasks. Furthermore, the discounted visitation measure is more desirable to learn than the stationary one, but may be challenging due to the exponentially decreasing influence of the time step at which states are visited \citep{islam2019marginalized}.

The main contribution of this paper is to introduce a MaxEntRL objective relying on a new intrinsic reward function for exploring effectively the state and action spaces, which also alleviates the previous limitations. This intrinsic reward function is the relative entropy of the discounted distribution of state-action features visited during future time steps. We prove two results motivating the MaxEntRL objective. First, this new objective is a lower bound on the MaxEntRL objective previously described, which integrates the marginal visitation distribution of states and actions. Second, the visitation distribution used in the new intrinsic reward function is the fixed point of a contraction operator. The intrinsic reward can therefore efficiently be computed off-policy using $N$-step state-action transitions and bootstrapping the operator. It is then possible to approximate the intrinsic reward function and learn a policy maximizing the extended rewards with existing RL algorithms. We demonstrate in our experiments how to adapt soft actor-critic \citep{haarnoja2018soft} to optimize the new objective, and we compare the effectiveness of exploration with different maximum entropy objectives. Experiments highlight improved visitation of features within individual trajectories, in exchange for slightly reduced visitation of features in expectation over different trajectories, as suggested by the lower bound. It also leads to improved convergence speed for learning exploration-only agents. Control performance remains similar across most methods on the considered benchmarks.

The visitation measure of future states and actions, which we use to extend the reward function in this article, has a well-established history in the development of RL algorithms. It was popularized by \citet{janner2020generative}, who learned the distribution of future states as a generalization of the successor features \citep{barreto2017successor}. They demonstrated that this distribution allows expressing the state-action value function by separating the influence of the dynamics and the reward function, and that it could be learned off-policy by exploiting its recursive expression. Several algorithms have been proposed to learn this distribution, using maximum likelihood estimation \citep{janner2020generative}, contrastive learning \citep{mazoure2023contrastive}, or diffusion models \citep{mazoure2023value}. These distributions of future states and actions have found applications in goal-based RL \citep{eysenbach2020c, eysenbach2022contrastive}, in offline pre-training with expert examples \citep{mazoure2023accelerating}, in model-based RL \citep{ma2023learning}, or in planning \citep{eysenbach2023contrastive}. We are the first to integrate them into the MaxEntRL framework for enhancing exploration. In parallel with our work, \citet{mohamed2025curiosity} developed a similar intrinsic reward dependent on the distribution of future states. 

The manuscript is organized as follows. In Section \ref{sec:prelim}, the RL problem and the MaxEntRL framework are formulated. In Section \ref{sec:maxentrl_visitation}, we introduce and discuss a new MaxEntRL objective. Section \ref{sec:off_pol_algo} details how to learn a model of the conditional state visitation probability that allows estimating this new objective. We finally present experimental results in Section \ref{sec:experiments} and conclude in Section \ref{sec:conclusion}.

\section{Background and Preliminaries} \label{sec:prelim}

\subsection{Markov Decision Processes}
This paper focuses on problems in which an agent makes sequential decisions in a stochastic environment \citep{sutton2018reinforcement}. The environment is modeled with an infinite-time Markov decision process (MDP) composed of a state space $\mathcal{S}$, an action space $\mathcal{A}$, an initial state distribution $p_0$, a transition distribution $p_1$, a bounded reward function $R$, and a discount factor $\gamma \in [0, 1)$. Agents interact in this MDP by providing actions sampled from a policy $\pi$. During this interaction, an initial state $s_0 \sim p_0(\cdot)$ is first sampled before the agent provides, at each time step $t$, an action $a_t \sim \pi(\cdot| s_t)$ leading to a new state $s_{t+1} \sim p_1(\cdot|s_t, a_t)$. In addition, after each action $a_t$ is executed, a reward $r_{t} = R(s_t, a_t) \in \mathbb{R}$ is observed. We denote the expected return of the policy $\pi$ by
\begin{align}
    J(\pi)
    &= \underset{
    \begin{subarray}{c}
    s_0  \sim p_0(\cdot) \\
    a_t \sim \pi(\cdot|s_t) \\
    s_{t+1} \sim p_1(\cdot| s_t, a_t) 
    \end{subarray}}{\mathbb{E}} \left [ \sum_{t=0}^\infty \gamma^t R(s_t, a_t) \right ] \, .
\end{align}
An optimal policy $\pi^*$ is one with maximum expected return
\begin{align}
    \pi^* \in \arg \max_\pi J(\pi) \, .
\end{align}

\subsection{Maximum Entropy Reinforcement Learning} \label{sec:intro_maxentrl}
In maximum entropy reinforcement learning (MaxEntRL), an optimal policy $\pi^*$ is approximated by maximizing a surrogate objective function $L(\pi)$, where the reward function from the MDP is extended by an intrinsic reward function. The latter is the (negated relative) entropy of some particular distribution. A general definition of the MaxEntRL objective function is
\begin{align}
    L(\pi)
    &= \underset{
    \begin{subarray}{c}
    s_0  \sim p_0(\cdot) \\
    a_t \sim \pi(\cdot|s_t) \\
    s_{t+1} \sim p_1(\cdot| s_t, a_t) 
    \end{subarray}}{\mathbb{E}} \hspace{-8pt} \left [ \sum_{t=0}^\infty \gamma^t \left ( R(s_t, a_t) + \lambda R^{int}(s_t, a_t) \right ) \right ] \, , \label{eq:objective_maxentrl}
\end{align}
where this objective depends on the intrinsic reward function $R^{int}$. We propose a generic formulation that, to the best of our knowledge, encompasses most existing MaxEntRL intrinsic rewards. Given a feature space $\mathcal{Z}$, a conditional feature distribution $q^\pi : \mathcal{S} \times \mathcal{A} \rightarrow \Delta(\mathcal{Z})$, depending on the policy $\pi$, and a relative measure $q^*\in \Delta(\mathcal{Z})$, the intrinsic reward function is
\begin{align}
    R^{int}(s, a)
    &= - KL_z \left [ q^\pi(z|s, a) \| q^*(z) \right ]
    = \underset{
    \begin{subarray}{c}
    z \sim q^\pi(\cdot|s, a)
    \end{subarray}}{\mathbb{E}} \left [ \log q^*(z) - \log q^\pi(z|s, a) \right ] \label{eq:intr_reward_maxentrl} \, .
\end{align}
Importantly, the intrinsic reward function is (implicitly) dependent on the policy $\pi$ through the distribution $q^\pi$. We define an optimal exploration policy as a policy that maximizes the expected sum of discounted intrinsic rewards only. Note that a policy maximizing $L(\pi)$ is generally not optimal, due to the potential gap between the optimum of the return $J(\pi)$ and the optimum of the learning objective $L(\pi)$. This subject is inherent to exploration with intrinsic rewards \citep{bolland2024behind}.

MaxEntRL algorithms optimize objective functions as defined in equation \eqref{eq:objective_maxentrl} depending on some intrinsic reward function that can be expressed as in equation \eqref{eq:intr_reward_maxentrl}. Each algorithm provides a particular estimate of the intrinsic reward and introduces a particular approach for maximizing the learning objective. Often, pseudo rewards are computed to extend the MDP rewards, which are jointly optimized, typically with (biased) policy gradient algorithms.

Many of the existing MaxEntRL algorithms optimize an objective that depends on the entropy of the policy for exploring the action space \citep{haarnoja2018soft, toussaint2009robot}. Features then correspond to actions such that the feature space is $\mathcal{Z} = \mathcal{A}$ and the conditional feature distribution is the policy, $q^\pi(z|s, a) = \pi(z| s)$ for all $a$. Other algorithms optimize objectives enhancing state space exploration \citep{hazan2019provably, lee2019efficient, islam2019marginalized, guo2021geometric}. The feature space is the state space $\mathcal{Z} = \mathcal{S}$. The conditional feature distribution $q^\pi(z|s, a)$ is either the marginal probability of states in trajectories of $T$ time steps, or the discounted state visitation measure, for all $s$ and $a$. In the literature, the relative measure $q^*(z)$ is usually a uniform distribution.

\section{MaxEntRL with Visitation Distributions} \label{sec:maxentrl_visitation}

\subsection{Definition of the MaxEntRL Objective} \label{sec:maxentrl_visitation_def}

In the following, we introduce a new MaxEntRL intrinsic reward based on the conditional state-action visitation probability $d^{\pi, \gamma}(\bar s, \bar a| s, a)$ and the conditional state visitation probability $d^{\pi, \gamma}(\bar s| s, a)$
\begin{align}
    d^{\pi, \gamma}(\bar s, \bar a| s, a) &= (1-\gamma) \pi(\bar a| \bar s) \sum_{\Delta=1}^\infty \gamma^{\Delta - 1} p_{\Delta}^\pi(\bar s| s, a) \label{eq:def_conditional_sa_visitation} \\
    d^{\pi, \gamma}(\bar s| s, a) &= (1-\gamma) \sum_{\Delta=1}^\infty \gamma^{\Delta - 1} p_{\Delta}^\pi(\bar s| s, a) \label{eq:def_conditional_s_visitation} \, ,
\end{align}
where $p_\Delta^\pi$ is the transition probability in $\Delta$ time steps with the policy $\pi$. The distribution from equation \eqref{eq:def_conditional_sa_visitation} can be factorized as a function of the distribution from equation \eqref{eq:def_conditional_s_visitation} such that $d^{\pi, \gamma}(\bar s, \bar a| s, a) = \pi(\bar a| \bar s) d^{\pi, \gamma}(\bar s| s, a)$. The conditional state (respectively, state-action) visitation probability distribution measures the states (respectively, states and actions) that are visited in expectation over infinite trajectories starting from a state and an action. Both distributions extend the (marginal discounted) state visitation/occupancy probability \citep{manne1960linear}, defined as $d^{\pi, \gamma}(\bar s) = (1-\gamma) \sum_{t=0}^{\infty} \gamma^{t} p^{\pi}_{t}(\bar s)$, where $p^{\pi}_{t}$ is the state probability at time $t$ under $p_0$ and $\pi$.

\begin{definition} \label{def:maxentrl_cv_def} Let us consider the feature space $\mathcal{Z}$ and the feature distribution $h: \mathcal{S} \times \mathcal{A} \rightarrow \Delta(\mathcal{Z})$. The intrinsic reward is defined by equation \eqref{eq:intr_reward_maxentrl}, for any relative measure $q^*$, with the distribution
\begin{align}
    q^\pi(z| s, a) &= \int h(z| \bar s, \bar a) d^{\pi, \gamma}(\bar s, \bar a| s, a) \: d\bar s \: d\bar a \, . \label{eq:def_q}
\end{align}
\end{definition}

Optimal exploration policies are here intrinsically motivated to take actions so that the discounted visitation measure of future features is distributed according to $q^*$ in each state and for each action. It allows selecting features that must be visited during trajectories according to prior knowledge about the problem if any. Alternatively, it allows exploring lower-dimensional feature spaces, or exploring some statistics from the state-action pairs.

Existing RL algorithms can be used to compute MaxEntRL policies following Definition \ref{def:maxentrl_cv_def} by computing for each state $s$ and action $a$ the additional (pseudo) reward
\begin{align}
    R^{int}(s, a) &= \log q^*(z) - \log q^\pi(z| s, a) \, , \label{eq:def_approx_r}
\end{align}
where $z \sim q^\pi(\cdot | s, a)$. This reward is a single-sample Monte-Carlo estimate of equation \eqref{eq:intr_reward_maxentrl}, unbiased for fixed $q^\pi$. This computation requires sampling features $z$ from the conditional distribution $q^\pi$ and estimating the probability of these samples $q^\pi(z| s, a)$. It can be achieved by solving the integral equation \eqref{eq:def_q}, e.g., numerically by sampling state-action pairs $(\bar s, \bar a) \sim d^{\pi, \gamma}(\cdot, \cdot| s, a)$ and features $z \sim h(\cdot| \bar s, \bar a)$, or learning a model of the conditional feature distribution. Section \ref{sec:off_pol_algo} provides a method for learning such a model off-policy.

\subsection{Relationship with Alternative MaxEntRL Objectives} \label{sec:comparison_maxentrl_obj}

Let us consider an MDP without rewards and study exploration policies. We assume for simplicity that the feature distribution is the identity mapping, so that $z = (\bar s, \bar a)$ and that $q^\pi(z | s, a) = d^{\pi, \gamma}(\bar s, \bar a | s, a) = d^{\pi, \gamma}(\bar s | s, a) \pi(\bar a | \bar s)$, according to Definition \ref{def:maxentrl_cv_def}. This MaxEntRL intrinsic reward function can be compared to one where the conditional distribution is $q^\pi(z | s, a) = d^{\pi, \gamma}(\bar s, \bar a) = d^{\pi, \gamma}(\bar s) \pi(\bar a | \bar s)$, for all $s$ and $a$, and where the MaxEntRL objective function simplifies to the entropy of that distribution. Theorem \ref{thr:lower_bound}, shown in Appendix \ref{apx:proof_lb}, relates the two objectives.

\begin{theorem} \label{thr:lower_bound}
For any policy $\pi$ and any relative measure $q^*$, the marginal and conditional visitation measures satisfy
\begin{multline*}
      \underset{
        \begin{subarray}{c}
            s, a \sim d^{\pi, \gamma}(\cdot , \cdot)
        \end{subarray}}{\mathbb{E}}  \left [
        -KL_{\bar s, \bar a} \left [ d^{\pi, \gamma}(\bar s, \bar a| s, a) || q^*(\bar s, \bar a) \right ] \right ] \\
        \leq - KL_{\bar s, \bar a}(d^{\pi, \gamma}(\bar s , \bar a) || q^*(\bar s , \bar a)) + L\, \sqrt{2\, KL_{\bar s, \bar a}(d^{\pi, \gamma}(\bar s , \bar a) || \tilde{d}^{\pi, \gamma}(\bar s , \bar a))} \, ,
\end{multline*}
where $L$ is a finite constant and $\tilde{d}^{\pi, \gamma}(\bar s, \bar a) = \mathbb{E}_{s, a \sim d^{\pi, \gamma}(\cdot, \cdot)} [ d^{\pi, \gamma}(\bar s, \bar a| s, a)]$.
\end{theorem}

Let us first assume that $\tilde{d}^{\pi, \gamma}(\bar s, \bar a) = d^{\pi, \gamma}(\bar s, \bar a)$. Then, the left-hand side corresponds to our new objective with $R(s,a) = 0$, which is a lower bound on the entropy of the marginal state-action visitation measure. If $d^{\pi, \gamma}(\bar s, \bar a | s, a) = q^*(\bar s, \bar a)$ almost everywhere, then the lower bound is zero, and $d^{\pi, \gamma}(\bar s, \bar a) = q^*(\bar s, \bar a)$ almost everywhere as well. It implies that our MaxEntRL objective is a lower bound on the alternative MaxEntRL objective with marginal visitation. Additionally, a policy maximizing our objective also maximizes the alternative objective, when achieving an intrinsic reward of zero. In the limit when the discount factor tends to one $\gamma \rightarrow 1$, this equivalence also holds as the effect of the initial state vanishes and both distributions converge to the stationary visitation distribution, provided it exists. In practice, the bound holds only when $\tilde{d}^{\pi, \gamma}(\bar s, \bar a)$ is close to $d^{\pi, \gamma}(\bar s, \bar a)$; it intuitively corresponds to a stationary assumption on the initial states.

This result connects MaxEntRL optimizing the entropy of the conditional visitation with MaxEntRL optimizing the entropy of the same distribution marginalized over initial states and actions. It can be straightforwardly extended to relate the conditional and marginal visitation of features, typically such that MaxEntRL algorithms optimizing intrinsic rewards from Definition \ref{def:maxentrl_cv_def} with $z = \bar s$ maximize lower bounds of MaxEntRL algorithms using the state visitation \citep{hazan2019provably}.

\section{Off-policy Learning of Conditional Visitation Models} \label{sec:off_pol_algo}

\subsection{Fixed-Point Properties of Conditional Visitation}

As explained in Section \ref{sec:maxentrl_visitation_def}, the MaxEntRL intrinsic reward function in Definition \ref{def:maxentrl_cv_def} can be computed by sampling from the conditional feature distribution and evaluating the probability of these samples. In this section, we establish useful properties of this conditional distribution.

Let us first recall that the conditional state-action visitation distribution accepts a recursive definition \citep{janner2020generative} that is a trivial fixed point of the operator $\mathcal{T}^\pi$ from Definition \ref{def:operator_T}.

\begin{definition} \label{def:operator_T} The operator $\mathcal{T}^\pi$ is defined over the space of conditional state-action distributions as
\begin{align*}
    \mathcal{T}^\pi q(\bar s, \bar a| s, a) 
    &= (1-\gamma) \pi(\bar a| \bar s) p_1(\bar s| s, a) + \gamma
    \underset{
        \begin{subarray}{c}
            s' \sim p_1(\cdot| s, a) \\
            a' \sim \pi(\cdot | s')
        \end{subarray}}{\mathbb{E}}  \left [ q(\bar s, \bar a| s', a')  \right ] \, .
\end{align*}
\end{definition}

Theorem \ref{thr:contraction_T} establishes that the operator $\mathcal{T}^\pi$ is a contraction mapping, which implies the uniqueness of its fixed point. Assuming the result of the operator could be computed (or estimated), the fixed point could also be computed by successive application of this operator. It would allow computing the conditional state-action visitation distribution, and later estimating the intrinsic reward function.

\begin{theorem} \label{thr:contraction_T} The operator $\mathcal{T}^\pi$ is $\gamma$-contractive in $\bar L_n$-norm, where $\bar L_n( f )^n = \sup_y \int | f(x| y) |^n \: dx$.
\end{theorem}

Definition \ref{def:operator_P} introduces the operator $\mathcal{P}^\pi$ that implicitly depends on the feature distribution $h$. If this distribution is the identity map, then both operators $\mathcal{P}^\pi$ and $\mathcal{T}^\pi$ are equal.

\begin{definition} \label{def:operator_P} The operator $\mathcal{P}^\pi$ is defined over the space of conditional feature distributions as
\begin{align*}
    \mathcal{P}^\pi q(z| s, a) 
    &= 
    \underset{
        \begin{subarray}{c}
            s' \sim p_1(\cdot| s, a) \\
            a' \sim \pi(\cdot | s')
        \end{subarray}}{\mathbb{E}}  \left [ (1-\gamma) h(z| s', a') + \gamma q(z| s', a') \right ] \, .
\end{align*}
\end{definition}

Next, we establish in Theorem \ref{thr:contraction_P} that this operator is also a contraction mapping, such that Theorem \ref{thr:contraction_T} can be considered a corollary of the present result. The fixed point could again theoretically be computed by successive application of this operator.

\begin{theorem} \label{thr:contraction_P} The operator $\mathcal{P}^\pi$ is $\gamma$-contractive in $\bar L_n$-norm, where $\bar L_n( f )^n = \sup_y \int | f(x| y) |^n \: dx$.
\end{theorem}

Finally, we establish in Theorem \ref{thr:fp_P} that the unique fixed point of $\mathcal{P}^\pi$ is the conditional distribution used in the MaxEntRL intrinsic reward from Definition \ref{def:maxentrl_cv_def}. Assuming we could approximate this fixed point, we would get a model to compute the reward as in equation \eqref{eq:def_approx_r}.

\begin{theorem} \label{thr:fp_P} The unique fixed point of the operator $\mathcal{P}^\pi$ is
\begin{align*}
    q^\pi(z| s, a) &= \int h(z| \bar s, \bar a) d^{\pi, \gamma}(\bar s, \bar a| s, a) \: d\bar s \: d\bar a \, .
\end{align*}
\end{theorem}

The theorems are shown in Appendix \ref{apx:proofs_operators}.

\subsection{TD Learning of Conditional Visitation Models} \label{sec:learning_visitation}

In practice, computing the result of the operator $\mathcal{P}^\pi$ (and $(\mathcal{P}^\pi)^N$ after $N$ applications) may be intractable when large state and action spaces are at hand or when these spaces are continuous. It furthermore requires having a model of the MDP. A common approach is then to rely on a function approximator $q_\psi$ to approximate the fixed point. Furthermore, similarly to TD-learning methods \citep{sutton2018reinforcement}, Theorem \ref{thr:contraction_P} suggests optimizing the parameters of this model $q_\psi$ to minimize the residual of the operator, measured with an $\bar L_n$-norm for which the operator is contractive. With this metric, estimating the residual would require estimating the probability of future features; it cannot be trivially minimized by stochastic gradient descent using transitions from the environment. Indeed, this residual involves the transition density, whose score would be required for gradient-based minimization but is not accessible in the model-free setting. We therefore propose to solve a surrogate minimum cross-entropy problem, in which stochastic gradient descent can be applied afterwards. For any policy $\pi$, the fixed point $q^\pi$ is approximated with a function approximator $q_\psi$ with parameter $\psi$ optimized to solve
\begin{align}
    \arg \min_\psi
    \underset{
        \begin{subarray}{c}
            s, a \sim \mu(\cdot, \cdot) \\
            \bar z \sim \left (\mathcal{P}^\pi \right )^N q_\psi(\cdot| s, a)
        \end{subarray}}{\mathbb{E}} \left [ - \log q_\psi(\bar z| s, a) \right ] \, , \label{eq:mle_optim_visitation_model}
\end{align}
where $\mu$ is an arbitrary distribution over the state and action space, and where $N$ is any positive integer. This optimization problem is related to minimizing the KL-divergence instead of an $\bar L_n$-norm \citep{bishop2006pattern}.

Let us make explicit how samples from the distribution $(\mathcal{P}^\pi )^N q_\psi(\bar z| s, a)$ can be generated from the MDP. By definition of the operator $\mathcal{P}^\pi$, the distribution $(\mathcal{P}^\pi )^N q_\psi(\bar z| s, a)$ is the mixture
\begin{align}
    (\mathcal{P}^\pi )^N q_\psi(\bar z| s, a)
    &= \left ( \sum_{\Delta=1}^{N} (1 - \gamma) \gamma^{\Delta - 1}
    \underset{
        \begin{subarray}{c}
            s' \sim p_\Delta^\pi(\cdot| s, a) \\
            a' \sim \pi(\cdot |s')
        \end{subarray}}{\mathbb{E}}
    \left [ h(\bar z| s', a') \right ] \right ) + \gamma^N
    \underset{
        \begin{subarray}{c}
            s' \sim p_N^\pi(\cdot| s, a) \\
            a' \sim \pi(\cdot |s')
        \end{subarray}}{\mathbb{E}} \left [ q_\psi(\bar z| s', a') \right ] \nonumber \\
	&= \sum_{\Delta = 1}^\infty \mathcal{G}_{1-\gamma}(\Delta) \underset{
        \begin{subarray}{c}
            s' \sim p_{\Delta'}^\pi(\cdot| s, a) \\
            a' \sim \pi(\cdot |s')
        \end{subarray}}{\mathbb{E}} \left [ b_\psi(\bar z| s', a', \Delta) \right ] \Bigg |_{\Delta' = \min(\Delta, N)} \, ,
\end{align}
where $\mathcal{G}_{1-\gamma}(\Delta)$ is the probability of $\Delta$ from a geometric distribution of parameter $1-\gamma$, and
\begin{align}
	b_\psi(\bar z|s, a, \Delta) = &
	\left\{ 
		\begin{array}{cl}
    	h(\bar z| s, a) & \Delta \leq N \\
    	q_\psi(\bar z| s, a) & \Delta > N
  		\end{array}
	\right.
\end{align}
Sampling from $(\mathcal{P}^\pi )^N q_\psi(\bar z| s, a)$ consists of sampling from the mixture.

Let us reformulate the problem in equation \eqref{eq:mle_optim_visitation_model} highlighting the elements from the mixture and applying importance weighting 
\begin{align}
	\arg \min_\psi
    &\underset{
        \begin{subarray}{c}
            s, a \sim \mu(\cdot, \cdot) \\
            \Delta \sim \mathcal{G}_{1-\gamma'}(\cdot) \\
            s' \sim p_{\Delta'}^\beta(\cdot| s, a) \\
            a' \sim \pi(\cdot |s') \\
            \bar z \sim b_\psi(\cdot|s', a', \Delta) 
        \end{subarray}}{\mathbb{E}} \left [ - \frac{\mathcal{G}_{1-\gamma}(\Delta)}{\mathcal{G}_{1-\gamma'}(\Delta)}\frac{p_{\Delta'}^\pi(s' | s, a)}{p_{\Delta'}^\beta(s' | s, a)} \log q_\psi(\bar z| s, a) \right ] \Bigg |_{\Delta' = \min(\Delta, N)} \, . \label{eq:mle_optim_visitation_model_v2}
\end{align}
Importance weighting is applied to the transition probability $p_{\Delta'}^\pi$ and to the geometric distribution $\mathcal{G}_{1-\gamma}$. It introduces the behavior policy $\beta$ and a pseudo discount factor $\gamma'$. The first allows off-policy learning, and the second helps prevent some elements of the mixture from having negligible probabilities, improving results in practice. When $\beta=\pi$ or when $N=1$, the second importance ratio simplifies to one; otherwise, it simplifies to a (finite) product of ratios of policies.

Learning $q_\psi$ from samples can be achieved by solving the problem in equation \eqref{eq:mle_optim_visitation_model_v2} as an intermediate step to existing RL algorithms. First, the objective function is estimated using transitions. Second, this estimate is differentiated, and the parameter $\psi$ is updated by gradient descent steps. In practice, the gradients generated by differentiating this loss function are biased. The influence of the parameter $\psi$ on the probability of the sample $\bar z$ is neglected when bootstrapping, i.e., the partial derivative of $\left (\mathcal{P}^\pi \right )^N q_\psi(\bar z| s, a)$ with respect to $\psi$ is neglected, and a target network is used. This is analogous to SARSA and TD-learning strategies \citep{sutton2018reinforcement}. Furthermore, we suggest neglecting the ratio of transition probabilities, which introduces a dependency of the distribution $q_\psi$ on the policy $\beta$. Finally, the model $q_\psi$ is used to compute the intrinsic rewards and update the policy.

\section{Experiments} \label{sec:experiments}

\subsection{Experimental Setting}
In this section, we detail the methodology applied to compare the MaxEntRL objectives discussed above. We use a suite of adapted environments for illustration, describe algorithms from the literature adapted to isolate the effect of the choice of objective from algorithmic considerations, and introduce measures to quantify the quality of exploration.

\paragraph{Environments.} Experiments are performed on environments from the Minigrid suite \citep{MinigridMiniworld23}. In the latter, an agent must travel across a grid containing walls and passages in order to reach a goal. The state space is a full observation of the maze, represented with an image, and the agent's orientation. The agent can take four different actions: turn left, turn right, move forward, or stand still. The need for exploration comes from the sparsity of the reward function, which is zero everywhere and equals one in the state to be reached.

\paragraph{Exploration strategies.} We compare three exploration strategies, i.e., three intrinsic reward functions and their corresponding MaxEntRL algorithm. The first exploration strategy motivates agents to perform actions uniformly. The feature space is the action space $\mathcal{Z} = \mathcal{A}$, the conditional feature distribution is the policy $q^\pi(z|s, a) = \pi(z| s)$ for all $a$, and the relative measure $q^*$ is uniform. The second exploration strategy motivates agents to visit uniformly the grid positions when starting from initial states. Here, the features $z \in \mathcal{Z}$ are the positions of the agent in the environment, the conditional distribution $q^\pi(z|s, a)$ is the discounted visitation measure of features, for all state $s$ and action $a$, and the relative measure $q^*$ is uniform. It corresponds to the alternative objective discussed in Section \ref{sec:comparison_maxentrl_obj}. The last exploration strategy is the one presented in Section \ref{sec:maxentrl_visitation_def}. Again, the features $z \in \mathcal{Z}$ are the positions of the agent in the environment, and the relative measure $q^*$ is uniform.

\paragraph{Algorithms.} Existing algorithms can be adapted to optimize the previous MaxEntRL objective by adding the intrinsic reward to the reward from the MDP during policy optimization. In some algorithms, as in ours, it requires learning an additional model of some visitation measure. In this paper we adapted soft actor-critic to incorporate the different intrinsic reward functions. We consider three variants, one for each exploration strategy. First, without additional intrinsic reward, it is already a MaxEntRL algorithm that enforces the entropy of the policy. Second, we adapt the algorithm from \citet{zhang2021exploration}, using SAC \citep{haarnoja2018soft} instead of PPO \citep{schulman2017proximal} to improve sample efficiency and using a categorical distribution rather than a variational auto-encoder to approximate the visitation measure, which is made possible as the state-action space is discrete. It allows optimizing the approximator without relying on the evidence lower bound. Third, we adapt SAC to incorporate our reward function as discussed previously and detailed in Appendix \ref{apx:algo}.

\paragraph{Exploration metrics.} The last step is to quantify the quality of the exploration policies, which most often consists in measuring the diversity of states (or features). Here we report two such metrics during learning. First, the entropy of features visited by policies
\begin{align}
	 - KL_z( d^{\pi, \gamma}(z) || q^*(z) ) \, ,
\end{align}
where $d^{\pi, \gamma}(z) = \mathbb{E}_{s, a \sim d^{\pi, \gamma}(\cdot, \cdot)} [ h(z| s, a) ]$.
Second, the entropy of features visited within a single trajectory, in expectation over trajectories,
\begin{align}
	 - \underset{
    \begin{subarray}{c}
    s_0  \sim p_0(\cdot) \\
    a_t \sim \pi(\cdot|s_t) \\
    s_{t+1} \sim p_1(\cdot| s_t, a_t) 
    \end{subarray}}{\mathbb{E}} \left [ KL_z(d^{\gamma}(z|\tau) || q^*(z) ) \right ] \, ,
\end{align}
where $\tau = (s_0, a_0, s_1, a_1, \dots)$ is the random trajectory, and where $d^{\gamma}(z|\tau) = (1-\gamma) \sum_{t=0}^{\infty} \gamma^t h(z| s_t, a_t)$ is the discounted distribution of features along this trajectory.
The first metric measures the diversity in expectation over episodes, whereas the second measures the diversity within individual episodes. The first metric also corresponds to the exploration objectives maximized by the second algorithm. Since $q^*$ is uniform over the grid positions, both metrics equal the entropy of the corresponding distribution up to the constant $\log |\mathcal{Z}|$; the following figures report these entropies. Both metrics are estimated every 500 iterations from 64 trajectories of 200 time steps, using the discounted empirical distribution of features over these steps.

\subsection{Experimental Results}
The methodology explained above is applied to various environments; further implementation details are given in Appendix \ref{apx:implementation}. Following the recommendations of \citet{agarwal2021deep}, all metrics are reported as interquartile mean (IQM) estimates with $95\%$ confidence intervals. The evolution of all metrics is shown in Appendix \ref{apx:learning}. We summarize the observations in this section.

\begin{figure}[t]
  \centering
  \includegraphics[width=0.9\linewidth]{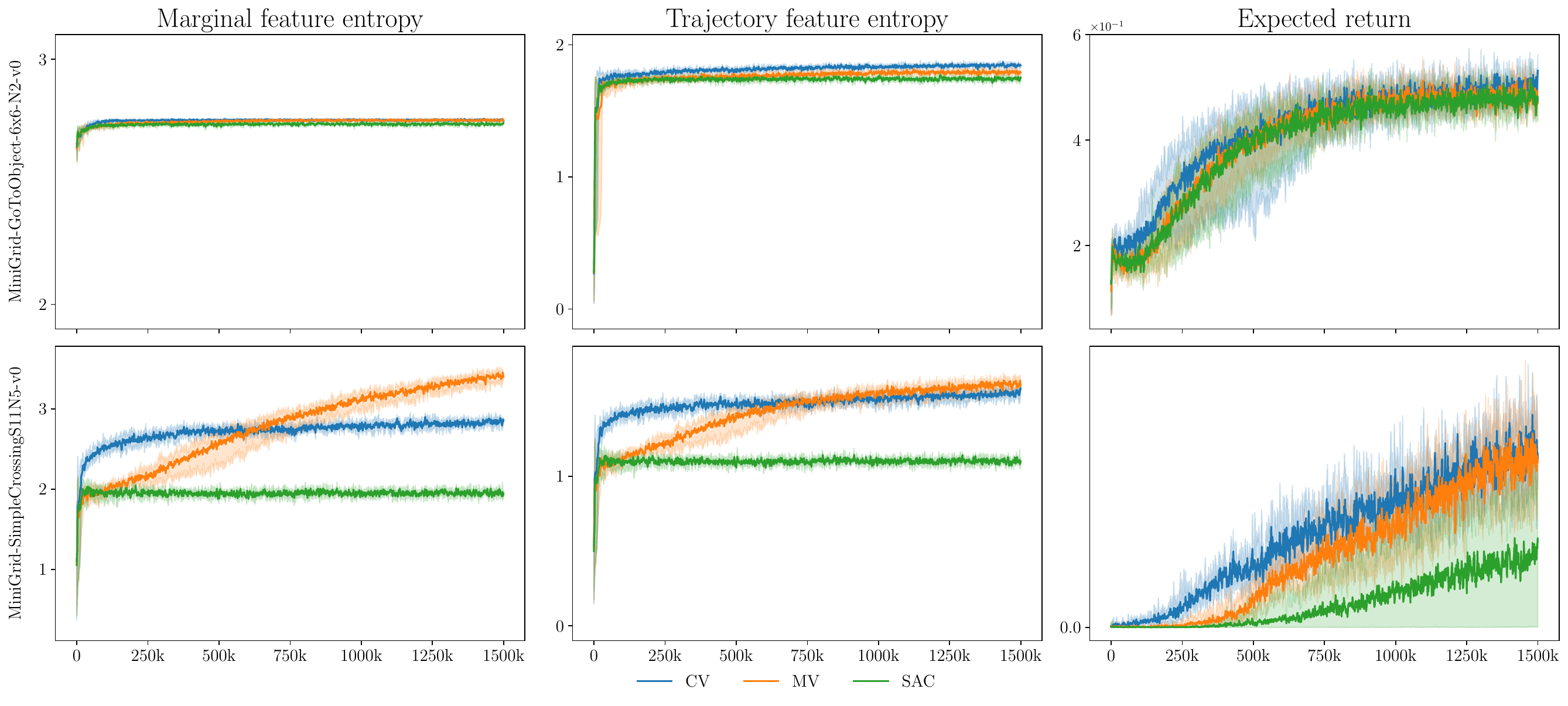}
  \caption{Learning results for two representative environments, selected from Appendix \ref{apx:learning}. The first two columns represent the evolution of the two exploration metrics defined in Section \ref{sec:experiments}, when learning exploration policies. The third column represents the evolution of the expected return when learning MaxEntRL control policies.}
  \label{fig:illustration}
\end{figure}

\paragraph{Marginal exploration.} Figure \ref{fig:learning_curve_marginal_entropy_exploration} in Appendix \ref{apx:learning} illustrates the evolution of the entropy of features visited by policies $- KL_z(d^{\pi, \gamma}(z) || q^*(z) )$ as a function of the learning iterations, when only optimizing the intrinsic rewards. We distinguish two situations. First, for some environments, the entropy does not evolve much during learning, and the three exploration strategies perform similarly. This is mostly due to the influence of the initial state distribution. In several environments, the initial position is drawn uniformly at random, such that the entropy of a random policy leads to high feature entropy due to symmetries. Finding optimal policies in such environments is arguably easy, as a large diversity of transitions will be observed without requiring complex MaxEntRL objectives. Second, for other environments, the entropy increases rapidly for the second algorithm, with marginal visitation measures (MV), and for the third algorithm, with conditional visitation measures (CV), and a high-entropy policy results from both optimizations. In these environments, MV achieves the highest entropy, followed closely by CV, while SAC performs poorly. It is worth noting that CV converges much faster than MV and eventually approaches the final entropy of MV despite optimizing a different objective, the remaining gap being small relative to their common gain over SAC. The first observation is likely a consequence of the off-policy estimation of the intrinsic reward function, which leads to improved sample efficiency, and the second may be justified by the bound from Theorem \ref{thr:lower_bound}. Here, the environments are apparently more complex to explore, and both advanced strategies allow observing a wide diversity of features in expectation. Figure \ref{fig:illustration} reports these results for two representative environments.

\paragraph{Episode exploration.} Figure \ref{fig:learning_curve_conditinal_entropy_exploration} in Appendix \ref{apx:learning} illustrates the evolution of the second metric as a function of the learning iterations, when only optimizing the intrinsic rewards. The metric improves as a function of the learning iterations in most cases. The best performance is most often achieved with the algorithm using CV, with SAC as the worst performer. The algorithm using MV nevertheless catches up with, and sometimes slightly exceeds, CV at the end of learning, notably in environments where the initial position is constant in each episode. Even in environments with uniform initial positions, motivating agents to explore leads to a larger variety of features visited during individual trajectories. Again, the intrinsic reward we present in this paper converges faster, which we attribute to the improved sample efficiency of the off-policy algorithm. The evolution of the entropy is reported for the same representative environments as before in Figure \ref{fig:illustration}.

\paragraph{Control policies.} The literature sometimes focuses on exploration only, but MaxEntRL usually aims to explore in order to eventually compute a high-performance policy. Figure \ref{fig:learning_curve_return_control} in Appendix \ref{apx:learning} reports the evolution of the expected return of policies when combining rewards from the environment with intrinsic rewards. In most environments, a sufficiently large weight to environment rewards combined with a large buffer leads to improving return during learning. In some environments, the complex exploration strategies allow learning faster, but no significant improvement was observed for the most favorable hyperparameters. The evolution of the expected return is reported for the same environments as before in Figure \ref{fig:illustration}.

\subsection{Discussion of Experiments}

Experiments highlight that complex MaxEntRL methods are sometimes necessary to achieve better feature space coverage. In particular, our method often allows better exploration of features within individual trajectories compared to more standard objectives. A likely justification is that our objective, relying on the conditional entropy, motivates agents to explore in the future, i.e., for the remainder of the trajectory, where alternative objectives are more influenced by the initial states and actions. To the best of our knowledge, it is unclear if there is a best metric to compare exploration strategies, and experiments seem to highlight that exploration is highly environment-dependent.

Several phenomena influence the learning of the visitation models. First, when $\gamma$ is close to one, the learning becomes unstable in practice. We hypothesize that it results from the increased importance of future states. Increasing the parameter $N$ helps mitigate the issue as there is less bootstrapping, reducing the risk of learning a biased target. Second, we neglect some importance weights in practice to reduce variance, which makes $q_\psi$ partially dependent on the behavior policy $\beta$. Bootstrapping still propagates long-term effects of the policy. In practice, regularizing the policy entropy even when using the two complex exploration objectives appeared to be critical to stabilize learning.

Finally, we relied on off-policy actor-critic for concreteness, yet the MaxEntRL objective is agnostic to the control backbone, and similar results should hold with other RL methods. We decided to share the same backbone to center the discussion around the MaxEntRL objective but acknowledge that the choice of algorithm may influence practical performance. Also, we observe that our method offers a practical alternative to directly maximizing marginal visitation, but we did not focus on potential theoretical advantages of different exploration objectives.

\section{Conclusion} \label{sec:conclusion}

In this paper, we presented a new MaxEntRL objective that provides intrinsic reward bonuses proportional to the entropy of the distribution of features built from the states and actions visited by the agent in future time steps. We show that the expected sum of these intrinsic rewards is a lower bound on an alternative maximum entropy objective that uses the marginal distribution of features starting from the initial state. The reward bonus can furthermore be estimated efficiently by sampling from the conditional distribution of visited features, which we proved to be the fixed point of a contraction mapping and which can be learned for any policy using batches of arbitrary transitions. This estimation procedure is easy to implement and can be integrated into existing RL algorithms by augmenting the reward with our intrinsic bonus. It was integrated into an end-to-end off-policy algorithm and benchmarked on several control problems. Experiments show improved feature visitation within individual trajectories in most environments and faster convergence for exploration-only agents, while control performance remains similar on the considered benchmarks.

This study reminds us that the exploration objective to pursue remains unclear and, in practice, depends on the environment. Future work includes clarifying this question theoretically and empirically. This includes studying the algorithm we presented and benchmarking it in more challenging environments. A key limitation of our algorithm is that the feature space is fixed a priori, and performance depends on this choice. A potential avenue is to learn it and explore maximum-information state-action feature spaces or reward-predictive feature spaces. Another minor practical limitation is that we estimate the conditional probabilities of future features with a parametric density model trained by minimizing cross-entropy, which restricts the model class, though it still covers mixtures of parametric distributions and normalizing flows. Using other models, such as latent space models or diffusion models, would require adapting the estimation and learning procedure.

\section*{Acknowledgment}
Adrien Bolland gratefully acknowledges the financial support of the \textit{Fonds de la Recherche Scientifique -- FNRS}. Gaspard Lambrechts is a Marie Skłodowska-Curie Actions (MSCA) postdoctoral fellow at McGill University, funded by Horizon Europe under grant 101284668. Part of this work was carried out while he was a postdoctoral fellow of the FNRS. He also gratefully acknowledges Wallonia-Brussels International for its support through a World Excellence Fellowship.

\appendix


\newpage
\bibliography{bibliography}
\bibliographystyle{rlj}

\beginSupplementaryMaterials

\section{Proof Lower Bound} \label{apx:proof_lb}

\begin{restatement}[\ref{thr:lower_bound}, restated]
For any policy $\pi$ and any relative measure $q^*$, the marginal and conditional visitation measures satisfy
\begin{multline*}
      \underset{
        \begin{subarray}{c}
            s, a \sim d^{\pi, \gamma}(\cdot , \cdot)
        \end{subarray}}{\mathbb{E}}  \left [
        -KL_{\bar s, \bar a} \left [ d^{\pi, \gamma}(\bar s, \bar a| s, a) || q^*(\bar s, \bar a) \right ] \right ] \\
        \leq - KL_{\bar s, \bar a}(d^{\pi, \gamma}(\bar s , \bar a) || q^*(\bar s , \bar a)) + L\, \sqrt{2\, KL_{\bar s, \bar a}(d^{\pi, \gamma}(\bar s , \bar a) || \tilde{d}^{\pi, \gamma}(\bar s , \bar a))} \, ,
\end{multline*}
where $L$ is a finite constant and $\tilde{d}^{\pi, \gamma}(\bar s, \bar a) = \mathbb{E}_{s, a \sim d^{\pi, \gamma}(\cdot, \cdot)} [ d^{\pi, \gamma}(\bar s, \bar a| s, a)]$.
\end{restatement}

\paragraph{Proof Theorem \ref{thr:lower_bound}} Let us assume that all distributions admit continuous, smooth, and strictly positive densities on their support with bounded log-ratios. Let $\tilde{d}^{\pi, \gamma}(\bar s, \bar a) = \mathbb{E}_{s, a \sim d^{\pi, \gamma}(\cdot, \cdot)} [ d^{\pi, \gamma}(\bar s, \bar a| s, a)]$. Let us develop using the convexity of KL divergence, positivity of KL divergence, integral properties and Pinsker's inequality
\begin{align*}
    & \underset{
        \begin{subarray}{c}
            s, a \sim d^{\pi, \gamma}(\cdot , \cdot)
        \end{subarray}}{\mathbb{E}}  \left [
        KL_{\bar s, \bar a} \left [ d^{\pi, \gamma}(\bar s, \bar a| s, a) || q^*(\bar s, \bar a) \right ] \right ] \\
        & \hspace{2cm} = \underset{
        \begin{subarray}{c}
            s, a \sim d^{\pi, \gamma}(\cdot , \cdot) \\
            \bar s, \bar a \sim d^{\pi, \gamma}(\cdot , \cdot| s, a)
        \end{subarray}}{\mathbb{E}}  \left [
        \log \frac{d^{\pi, \gamma}(\bar s, \bar a| s, a)}{q^*(\bar s, \bar a)} \right ] \\
		& \hspace{2cm} = \underset{
        \begin{subarray}{c}
            s, a \sim d^{\pi, \gamma}(\cdot , \cdot) \\
            \bar s, \bar a \sim d^{\pi, \gamma}(\cdot , \cdot| s, a)
        \end{subarray}}{\mathbb{E}}  \left [
        \log \frac{d^{\pi, \gamma}(\bar s, \bar a| s, a)}{d^{\pi, \gamma}(\bar s , \bar a)} + \log \frac{d^{\pi, \gamma}(\bar s , \bar a)}{q^*(\bar s, \bar a)} \right ] \\
        & \hspace{2cm} \geq \underset{
        \begin{subarray}{c}
            s, a \sim d^{\pi, \gamma}(\cdot , \cdot) \\
            \bar s, \bar a \sim d^{\pi, \gamma}(\cdot , \cdot| s, a)
        \end{subarray}}{\mathbb{E}}  \left [
        \log \frac{d^{\pi, \gamma}(\bar s , \bar a)}{q^*(\bar s, \bar a)} \right ] \\
        & \hspace{2cm} = \int \tilde{d}^{\pi, \gamma}(\bar s, \bar a) \log \frac{d^{\pi, \gamma}(\bar s , \bar a)}{q^*(\bar s, \bar a)} \: d\bar s \: d\bar a \\
        & \hspace{2cm} = \int d^{\pi, \gamma} \log \frac{d^{\pi, \gamma}}{q^*} \: d\bar s \: d\bar a - \int (d^{\pi, \gamma} - \tilde{d}^{\pi, \gamma}) \log \frac{d^{\pi, \gamma}}{q^*} \: d\bar s \: d\bar a \\
        & \hspace{2cm} = KL_{\bar s, \bar a}(d^{\pi, \gamma}(\bar s , \bar a)  || q^*(\bar s , \bar a)) - \int (d^{\pi, \gamma} - \tilde{d}^{\pi, \gamma}) \log \frac{d^{\pi, \gamma}}{q^*} \: d\bar s \: d\bar a \\
        & \hspace{2cm} \geq KL_{\bar s, \bar a}(d^{\pi, \gamma}(\bar s , \bar a) || q^*(\bar s , \bar a)) - \left | \int (d^{\pi, \gamma} - \tilde{d}^{\pi, \gamma}) \log \frac{d^{\pi, \gamma}}{q^*} \: d\bar s \: d\bar a \right | \\
        & \hspace{2cm} \geq KL_{\bar s, \bar a}(d^{\pi, \gamma}(\bar s , \bar a) || q^*(\bar s , \bar a)) - L \int \left | d^{\pi, \gamma} - \tilde{d}^{\pi, \gamma} \right | \: d\bar s \: d\bar a \\
        & \hspace{2cm} \geq KL_{\bar s, \bar a}(d^{\pi, \gamma}(\bar s , \bar a) || q^*(\bar s , \bar a)) - 2\, L\, TV( d^{\pi, \gamma}, \tilde{d}^{\pi, \gamma}) \\
        & \hspace{2cm} \geq KL_{\bar s, \bar a}(d^{\pi, \gamma}(\bar s , \bar a) || q^*(\bar s , \bar a)) - L\, \sqrt{2\, KL_{\bar s, \bar a}(d^{\pi, \gamma}(\bar s , \bar a) || \tilde{d}^{\pi, \gamma}(\bar s , \bar a))}
\end{align*}
\hfill $\square$

\newpage
\section{Proofs Theorems on Contractions} \label{apx:proofs_operators} 

\begin{restatement}[\ref{thr:contraction_T}, restated]
The operator $\mathcal{T}^\pi$ is $\gamma$-contractive in $\bar L_n$-norm, where $\bar L_n( f )^n = \sup_y \int | f(x| y) |^n \: dx$.
\end{restatement}

\paragraph{Proof Theorem \ref{thr:contraction_T}.} This theorem is a particular case of Theorem \ref{thr:contraction_P}, where the feature space $\mathcal{Z} = \mathcal{S} \times \mathcal{A}$ and the mapping $h$ is a Dirac
\begin{align*}
    h(z| s, a) = \delta_{(s, a)}(z) \, .
\end{align*}
\hfill $\square$

\begin{restatement}[\ref{thr:contraction_P}, restated]
The operator $\mathcal{P}^\pi$ is $\gamma$-contractive in $\bar L_n$-norm, where $\bar L_n( f )^n = \sup_y \int | f(x| y) |^n \: dx$.
\end{restatement}

\paragraph{Proof Theorem \ref{thr:contraction_P}.} For all conditional distributions $f$ and $g$
\begin{align*}
    \sup_{s, a} L_n(\mathcal{P}^\pi f(\cdot| s, a), \mathcal{P}^\pi g(\cdot| s, a))^n
    &= \sup_{s, a} \int | \mathcal{P}^\pi f(\bar z| s, a) - \mathcal{P}^\pi g(\bar z| s, a) |^n \: d \bar z \\
    &= \gamma^n \sup_{s, a} \int \left |
    \underset{
        \begin{subarray}{c}
            s' \sim p_1(\cdot| s, a) \\
            a' \sim \pi(\cdot | s')
        \end{subarray}}{\mathbb{E}}  \left [ f(\bar z| s', a') - g(\bar z| s', a') \right ]  \right |^n \: d \bar z \\
    &\leq \gamma^n \sup_{s, a} \int
    \underset{
        \begin{subarray}{c}
            s' \sim p_1(\cdot| s, a) \\
            a' \sim \pi(\cdot | s')
        \end{subarray}}{\mathbb{E}}  \left [  | f(\bar z| s', a') - g(\bar z| s', a') |^n\right ] \: d \bar z \\
    &= \gamma^n \sup_{s, a}
    \underset{
        \begin{subarray}{c}
            s' \sim p_1(\cdot| s, a) \\
            a' \sim \pi(\cdot | s')
        \end{subarray}}{\mathbb{E}}  \left [ \int | f(\bar z| s', a') - g(\bar z| s', a') |^n \: d \bar z \right ] \\
    &\leq \gamma^n \sup_{s, a} \sup_{s', a'} \left (  \int | f(\bar z| s', a') - g(\bar z| s', a') |^n \: d \bar z \right ) \\
    &= \gamma^n \sup_{s', a'}  \int | f(\bar z| s', a') - g(\bar z| s', a') |^n \: d \bar z \\
    &= \gamma^n \sup_{s, a}  L_n(f(\cdot| s, a), g(\cdot| s, a))^n
\end{align*}
\hfill $\square$

\newpage
\begin{restatement}[\ref{thr:fp_P}, restated]
The unique fixed point of the operator $\mathcal{P}^\pi$ is
\begin{align*}
    q^\pi(z| s, a) &= \int h(z| \bar s, \bar a) d^{\pi, \gamma}(\bar s, \bar a| s, a) \: d\bar s \: d\bar a \, .
\end{align*}
\end{restatement}

\paragraph{Proof Theorem \ref{thr:fp_P}.} The operator $\mathcal{P}^\pi$ is contractive, which implies that there exists a unique fixed point. Let us apply the operator $\mathcal{P}^\pi$ to the distribution $q^\pi$
\begin{align*}
    \mathcal{P}^\pi q^\pi(z|s, a)
    &= \mathcal{P}^\pi \int h(z|\bar s, \bar a) d^{\pi, \gamma}(\bar s, \bar a| s, a) \: d\bar s \: d\bar a \\
    &= \underset{
        \begin{subarray}{c}
            s' \sim p_1(\cdot| s, a) \\
            a' \sim \pi(\cdot | s')
        \end{subarray}}{\mathbb{E}}  \left [ (1-\gamma) h(z| s', a') + \gamma \left ( \int h(z|\bar s, \bar a) d^{\pi, \gamma}(\bar s, \bar a| s', a') \: d\bar s \: d\bar a \right ) \right ] \\
	&= \int h(z|\bar s, \bar a) \left ( \underset{
        \begin{subarray}{c}
            s' \sim p_1(\cdot| s, a) \\
            a' \sim \pi(\cdot | s')
        \end{subarray}}{\mathbb{E}}  \left [ (1-\gamma) \delta_{\bar s, \bar a}(s', a') + \gamma d^{\pi, \gamma}(\bar s, \bar a| s', a')\right ] \right ) \: d\bar s \: d\bar a \\
	&= \int h(z|\bar s, \bar a) \left ( (1-\gamma) \pi(\bar a |\bar s) p_1(\bar s |s, a)
	+ \gamma \underset{
        \begin{subarray}{c}
            s' \sim p_1(\cdot| s, a) \\
            a' \sim \pi(\cdot | s')
        \end{subarray}}{\mathbb{E}}  \left [d^{\pi, \gamma}(\bar s, \bar a| s', a')\right ] \right ) \: d\bar s \: d\bar a \\
	&= \int h(z|\bar s, \bar a) d^{\pi, \gamma}(\bar s, \bar a| s, a) \: d\bar s \: d\bar a \\
	&= q^\pi(z|s, a) \, .
\end{align*}
\hfill $\square$

\newpage
\section{Soft Actor-Critic with Conditional Visitation Measure} \label{apx:algo}

In the following, we adapt soft actor-critic \citep{haarnoja2018soft}, itself an adaptation of off-policy actor-critic \citep{degris2012off}, to include additional intrinsic rewards. In essence, soft actor-critic estimates the state-action value function with a parameterized critic $Q_\phi$, which is learned using expected SARSA (sometimes called generalized SARSA), and updates the parameterized policy $\pi_\theta$ with approximate policy iteration (i.e., off-policy policy gradient), all based on one-step transitions stored in a replay buffer $\mathcal{D}$. The actor and critic loss functions are furthermore extended with the log-likelihood of actions weighted by the parameter $\lambda_{SAC}$; it is therefore called soft and considered a MaxEntRL algorithm regularizing the entropy of policies. In the particular case in which $\lambda_{SAC}$ equals zero, the algorithm boils down to a slightly revisited implementation of off-policy actor-critic. 

Soft actor-critic is adapted to MaxEntRL with the intrinsic reward function defined in Section \ref{sec:maxentrl_visitation_def}, as follows. First, $N$-step transitions are stored in the buffer $\mathcal{D}$ instead of one-step transitions. Second, the conditional feature distribution is estimated with a function approximator $q_\psi$ and learned with stochastic gradient descent. Third, at each iteration of the critic updates, the reward provided by the MDP is extended with the intrinsic reward. Algorithm \ref{algo:offpol_exploration} summarizes the learning steps.

Formally, the parameterized critic $Q_\phi$ is iteratively updated by performing stochastic gradient descent steps on the loss function
\begin{align}
    \mathcal{L}(\phi) &=
    \underset{
        \begin{subarray}{c}
            s_t, a_t \sim \mathcal{D}
        \end{subarray}}{\mathbb{E}}  \left [ \left ( Q_\phi(s_t, a_t) - y \right )^2 \right ] \label{eq:loss_critic} \\
	y &= \lambda_R R(s_t, a_t) + \lambda R^{int}(s_t, a_t) + \gamma \left ( Q_{\phi'}(s_{t+1}, a_{t+1'}) - \lambda_{SAC} \log \pi_\theta(a_{t+1'}| s_{t+1}) \right ) \, ,
\end{align}
where $a_{t+1'} \sim \pi_\theta(\cdot| s_{t+1})$, and where $\phi'$ is the target network parameter. 

Furthermore, the policy $\pi_\theta$ is updated by performing gradient descent steps on the loss function
\begin{align}
    \mathcal{L}(\theta) &= -
    \underset{
        \begin{subarray}{c}
            s_t, a_t \sim \mathcal{D}
        \end{subarray}}{\mathbb{E}}  \left [ \log \pi_\theta(a_{t'}| s_t) A(s_t, a_{t'}) \right ] \label{eq:loss_actor}\\
        A(s_t, a_{t'}) &= Q_\phi(s_t, a_{t'}) - \lambda_{SAC} \log \pi_\theta(a_{t'}| s_t) \, ,
\end{align}
where $a_{t'} \sim \pi_\theta(\cdot| s_t)$.

\begin{algorithm}
\caption{SAC with conditional visitation measure for exploration}
\label{algo:offpol_exploration}
\begin{algorithmic}
\STATE Initialize the policy $\pi_\theta$, the soft critic $Q_\phi$, and the conditional feature model $q_\psi$
\STATE Initialize the critic target $Q_{\phi'}$ and visitation target $q_{\psi'}$
\STATE Initialize the replay buffer with $N$-step transitions
\WHILE{Learning}
    \STATE Sample transitions from the policy $\pi_\theta$ and add them to the buffer
    \WHILE{Update the visitation model}
    	\STATE Sample a batch of $N$-step transitions from the buffer
        \STATE Update the model $q_\psi$
    \ENDWHILE
    \WHILE{Update the critic}
    	\STATE Sample a batch of $N$-step transitions from the buffer (use only the 1-step transitions)
    	\STATE For each element of the batch sample $z_t \sim q_\psi(\cdot| s_t, a_t)$
    	\STATE Estimate the intrinsic reward $R^{int}(s_t, a_t) = \log q^*(z_t) - \log q_\psi(z_t|s_t, a_t)$
    	\STATE Perform a stochastic gradient descent step on $\mathcal{L}(\phi)$
    \ENDWHILE
    \STATE Sample a batch of $N$-step transitions from the buffer (use only the 1-step transitions)
    \STATE Perform a stochastic gradient descent step on $\mathcal{L}(\theta)$
    \STATE Update the target parameters with Polyak averaging
\ENDWHILE
\end{algorithmic}
\end{algorithm}

\newpage
\section{Implementation Details} \label{apx:implementation}
This section aims to clarify the practical implementation and how it differs in practice from the simplified algorithms presented.

The agent observes an image that is processed by a convolutional neural network into a state feature. The policy $\pi_\theta$ is a forward network that processes this feature and that outputs a categorical distribution over the action representation. The critic $Q_\phi$ is a neural network that averages the state feature with a projection of the action representation, computed by a forward network, and processes the result with a final forward network that outputs a scalar. In CV, the distribution model $q_\psi$ is also a neural network that takes the same input as the critic $Q_\phi$ and outputs a categorical distribution over a one-hot-encoding representation of positions. In MV, the visitation distribution model $q_\psi$ is a marginal distribution over the same one-hot-encoding representation. This marginal model is learned from transitions of the replay buffer, weighted by $(1-\gamma)\gamma^{t}$ where $t$ is the time step of the transition within its episode, without importance correction for the behavior policy; the model thus estimates the discounted feature visitation of the buffer distribution rather than that of the current policy, which we found more stable in practice.

The implementation of soft actor-critic differs slightly from that of the original work \citep{haarnoja2018soft}. The actor loss equation is based on the log-trick instead of the reparameterization trick, the expected SARSA update of the critic is approximated by sampling, and a single value function is learned, as implemented in CleanRL \citep{huang2022cleanrl}.

In CV, the feature $z_t$ used to estimate the intrinsic reward in equation \eqref{eq:def_approx_r} is furthermore not sampled from the model $q_\psi$ directly. It is instead sampled from the bootstrapped $N$-step mixture serving as target for learning the density in Section \ref{sec:learning_visitation}: a horizon is drawn from the geometric distribution with pseudo discount factor $\gamma'$, the observed feature from the stored $N$-step transition is used when the horizon is at most $N$, and a sample from the bootstrapped model is used otherwise. Target networks are used for the density model and for the policy providing the future action, both when bootstrapping and when evaluating the log-probability of the sample, and no importance weights are applied. Relying on observed features rather than model samples reduces the bias introduced by an imperfectly learned density model. In addition, sampling horizons with $\gamma' < \gamma$ shifts the reward toward features visited in the near future; in practice, it speeds up the initial increase of the entropy while leaving the final entropies essentially unchanged.

Some last practical choices were made and should be noted. The replay buffer is first filled with transitions collected with the randomly initialized policy network; transitions are then replaced following a FIFO policy during learning iterations. The number of transitions sampled per learning iteration is a hyperparameter. The actor update uses replay time step weights of the form $\gamma^{t-t_{\min}}$ and relies on an optional centering of the soft $Q$-values, which we refer to as actor centering. During the critic updates, the intrinsic reward is bounded on both sides: a small probability $\epsilon = 10^{-4}$ is added to the model density before taking its logarithm, which caps the maximum intrinsic reward when densities vanish, and the reward is clipped from below at $\log 10^{-2}$ to avoid negative reward overshoots when densities are too concentrated.

The parameters of all previous models are updated independently using the Adam optimization rule \citep{kingma2014adam}.

Hyperparameters were selected to maximize the area under the learning curve of the marginal feature entropy in the exploration experiments. A grid search was performed over the learning rates, the Polyak averaging weights were reduced when instabilities were observed, and the remaining parameters were fixed to standard values. Configurations leading to large instabilities were rejected.

Table \ref{tab:parameters} summarizes the hyperparameters used for the experiments.

\begin{table}[p]
\centering
\caption{Hyperparameters}
\label{tab:parameters}
\begin{tabular}{l|c}
    Parameter                               & Value \\
    \hline
    Convolutional layer                     & $2$ \\
    Neurons for each convolutional layer    & $32$ \\
    State-feature size                      & $256$ \\
    Forward layers policy                   & $2$ \\
    Forward layers critic                   & $2$  \\
    Neurons forward layers                  & $256$ \\
    Action-projection size critic           & $256$ \\
    Learning rate policy                    & $10^{-6}$ \\
    Learning rate critic                    & $10^{-6}$ \\
    Maximum trajectory length               & $200$ \\
    Buffer size                             & $100000$ \\
    Initial buffer fill (initial policy)    & $10000$ \\
    Batch size                              & $128$ \\
    Critic target update weight $\tau$      & $0.005$ \\
    Discount factor $\gamma$                & $0.98$ \\
    Extrinsic reward weight (control experiments) & $5$ \\
    \hline
    SAC $\lambda_{SAC}$                     & $0.01$ \\
    \hline
    Forward layers visitation model         & $2$  \\
    Neurons forward layers                  & $256$ \\
    Action-projection size visitation model & $256$ \\
    Learning rate visitation model          & $10^{-6}$ \\
    MaxEntRL $\lambda$                      & $0.005$ \\
    Density model target update weight $\tau$ & $0.005$ \\
    Bootstrapping steps $N$ (CV)            & $8$ \\
    Pseudo discount factor $\gamma'$ (CV)   & $0.95$ \\
    Policy target update weight $\tau$ (CV) & $0.005$ \\
\end{tabular}
\end{table}

\newpage
\section{Experimental Results} \label{apx:learning}

\begin{figure}[H]
  \centering
  \includegraphics[width=0.95\linewidth]{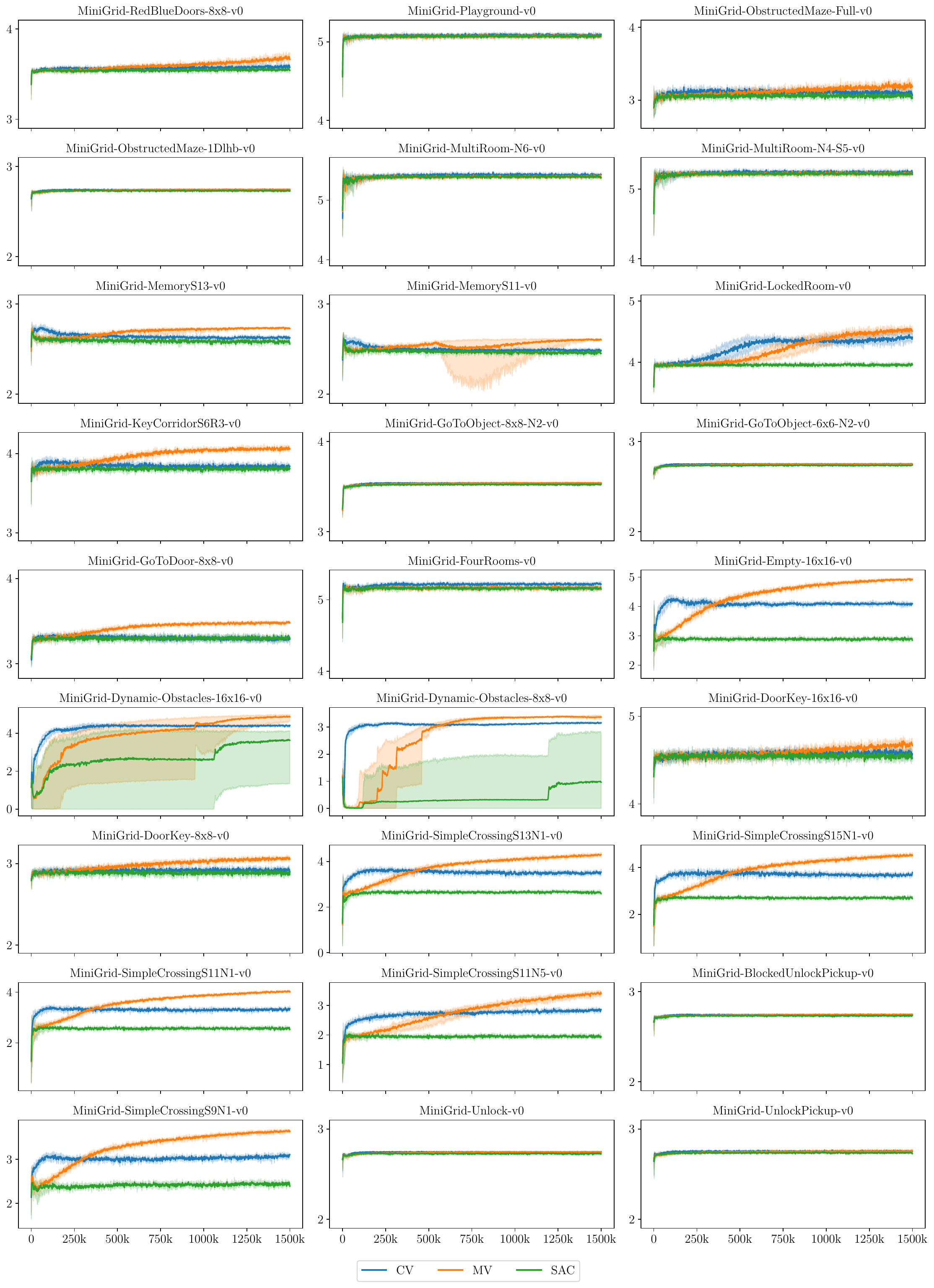}
  \caption{Evolution of the entropy of the discounted visitation probability measure of the position of the agent on the grid when computing exploration policies (i.e., when neglecting the rewards of the MDP). The entropy is computed empirically with Monte Carlo simulations. For each iteration, the interquartile mean over 5 runs is reported, along with its $95\%$ confidence interval.}
  \label{fig:learning_curve_marginal_entropy_exploration}
\end{figure}

\begin{figure}[H]
  \centering
  \includegraphics[width=0.95\linewidth]{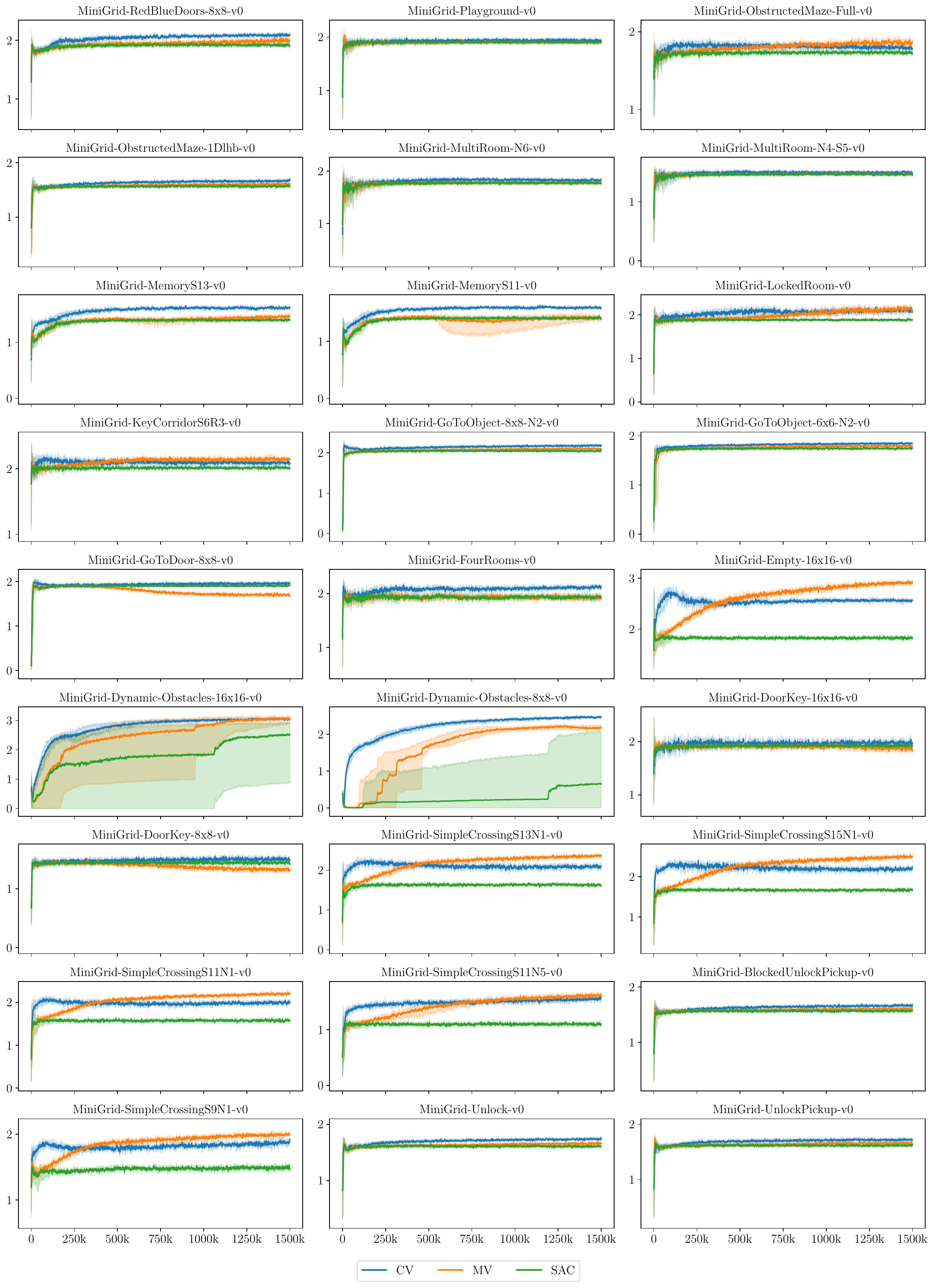}
  \caption{Evolution of the entropy of the discounted visitation probability measure of the position of the agent on the grid within single trajectories, averaged over trajectories, when computing exploration policies (i.e., when neglecting the rewards of the MDP). The entropy is computed empirically with Monte Carlo simulations. For each iteration, the interquartile mean over 5 runs is reported, along with its $95\%$ confidence interval.}
  \label{fig:learning_curve_conditinal_entropy_exploration}
\end{figure}

\begin{figure}[H]
  \centering
  \includegraphics[width=0.95\linewidth]{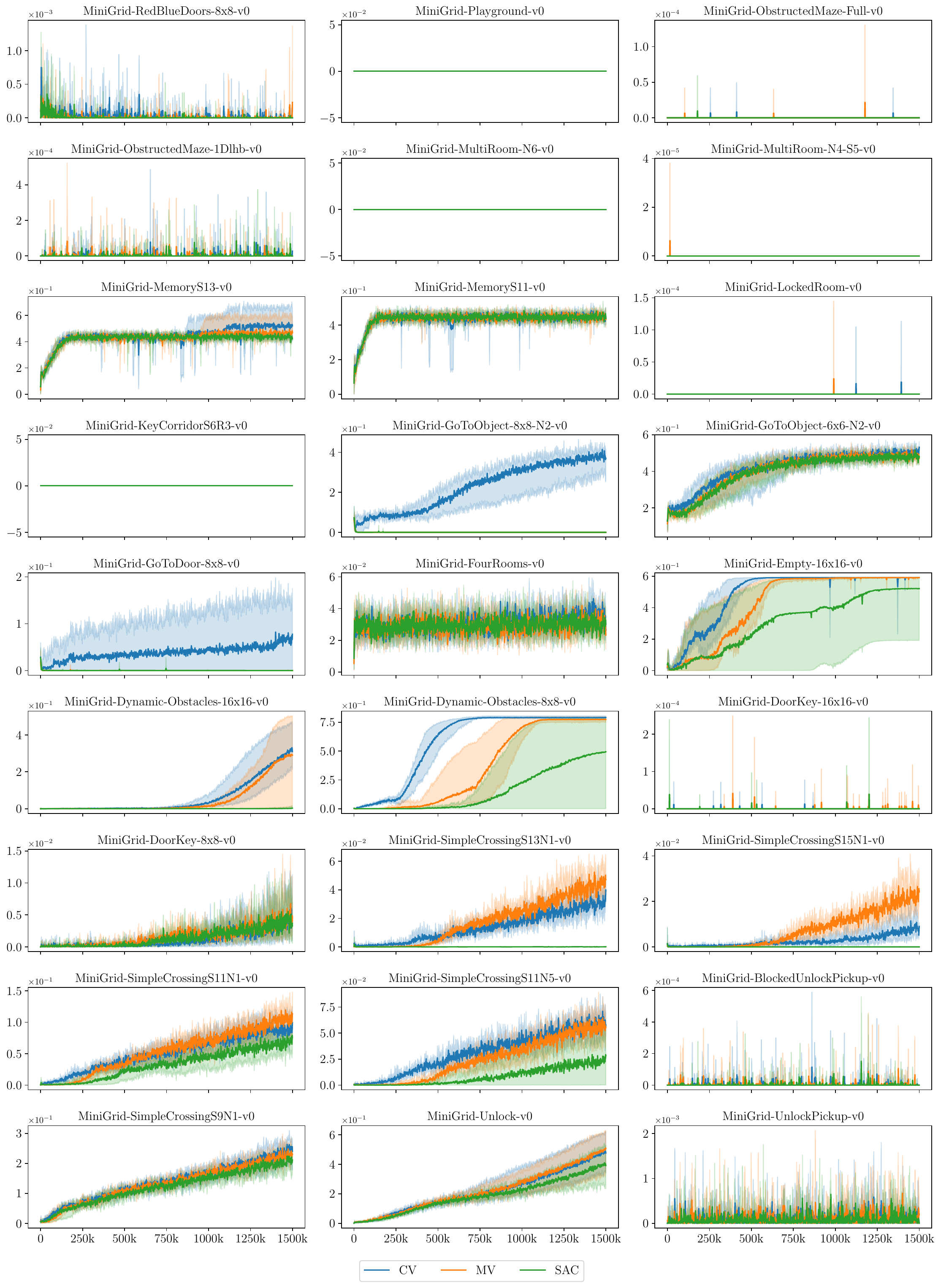}
  \caption{Expected return during the policy optimization. The expectation is computed empirically with Monte Carlo simulations. For each iteration, the interquartile mean over 5 runs is reported, along with its $95\%$ confidence interval.}
  \label{fig:learning_curve_return_control}
\end{figure}

\end{document}